# ADAPTIVE TRANSIT SIGNAL PRIORITY BASED ON DEEP REINFORCEMENT LEARNING AND CONNECTED VEHICLES IN A TRAFFIC MICROSIMULATION ENVIRONMENT

Dickness Kakitahi Kwesiga[1], Angshuman Guin[1], Michael Hunter[1]

[1]School of Civil and Environmental Engineering, Georgia Institute of Technology, Atlanta, GA, USA

**ABSTRACT**

Model free reinforcement learning (RL) provides a potential alternative to earlier formulations of adaptive transit signal priority (TSP) algorithms based on mathematical programming that require complex and nonlinear objective functions. This study extends RL-based traffic control to include TSP. Using a microscopic simulation environment and connected vehicle data, the study develops and tests a TSP event-based RL agent that assumes control from another developed RL-based general traffic signal controller. The TSP agent assumes control when transit buses enter the dedicated short-range communication (DSRC) zone of the intersection. This agent is shown to reduce the bus travel time by about 21%, with marginal impacts to general traffic at a saturation rate of 0.95. The TSP agent also shows slightly better bus travel time compared to actuated signal control with TSP. The architecture of the agent and simulation is selected considering the need to improve simulation run time efficiency.

## 1 INTRODUCTION

Transit signal priority research is mainly focused on developing optimization strategies to improve bus performance while limiting impacts to the general traffic. Using traffic microscopic simulation environments, several studies have tested adaptive transit signal priority (TSP) algorithms that incorporate connected vehicle (CV) data. Data from CVs enables the formulation of more precise actuation strategies and better evaluation of traffic states (Cvijovic et al. 2022; Mohammadi et al. 2020; Wang et al. 2020).

Studies such as Li et al. (2011) attempt to formulate signal timing optimization algorithms for TSP based on mathematical programming. However, these study's complex and nonlinear objective functions require high computational resources. Model free RL approaches may be a suitable alternative requiring reduced resources. Several previous studies have developed RL-based traffic control algorithms in simulation environments and shown that under the right conditions these algorithms can outperform conventional fixed and actuated signal timing plans (Bouktif et al. 2023; Li et al. 2020; Liu et al. 2022). Only a few recent studies have attempted to extend the RL-based signal control algorithms to include TSP by modifying state and reward function definitions (Cheng et al. 2022; Hu et al. 2023; Long et al. 2022; Shen et al. 2023; Yang and Fan 2024; Zhong et al. 2023). Compared to general traffic, bus arrivals at intersections are often sparse occurrences which limits the number of available samples for RL algorithms to learn TSP control. The majority of RL-based TSP studies inflate bus arrivals in simulation to generate sufficient training samples. While the resulting algorithms may be optimal at high bus frequencies, non-priority movements may be unfairly penalized at low bus frequencies.

To advance TSP this study uses a microscopic simulation environment to develop and train two traffic signal control RL agents. The first agent controls the intersection under general traffic conditions, i.e., in the absence of a bus. The second agent is event based, triggered to assume intersection signal control when a bus enters the connected vehicle (CV) dedicated short-range communication (DSRC) zone of the intersection, providing TSP. In testing the algorithm all buses are allowed to trigger the TSP agent but logical conditions can easily be added to limit TSP service to only buses meeting certain criteria, for example buses behind schedule or above a set occupancy. In addition to the RL agent development, the serialized online training in the simulation environment required improved run time efficiencies. The architecture of the algorithm is selected considering this need to improve simulation run time efficiency.

## 2 RELATED WORK

### 2.1 RL Traffic Signal Controllers

Several previous studies have developed RL-based traffic control algorithms. These studies have shown that RL-based signal control can potentially be superior to conventional fixed-time and actuated signal control. The control decisions/actions mainly involve deciding the next phase in a fixed or variable phasing sequence (Bouktif et al. 2023; Li et al. 2020; Li et al. 2016; Liu et al. 2022) and deciding the green duration of the next phase in fixed phasing sequence (Aslani et al. 2019; Bálint et al. 2022; Casas 2017; Lee et al. 2022; Li et al. 2021; Shabestary et al. 2020). Most of the studies have used deep q-network (DQN) and its variations, including double deep Q-networks (DDQN), dueling double deep Q-networks (3DQN), extended Dueling Double Deep Q-learning (e3DQN), DQN with prioritized experience replay, etc. Other studies have used actor critic (AC) based algorithms including advantage actor critic (A2C), double deep policy gradients (DDPG), and proximal policy optimization (PPO). Traffic states and rewards are defined/formulated using data that is, or can be, available from traffic sensors in the network and signal controllers, and more recently from CVs.

### 2.2 Traffic Simulation Environments

The reviewed studies have used microscopic, mesoscopic, and macroscopic simulation engines in developing and testing RL models. In recent efforts, microscopic approaches are most common as they provide detailed vehicle movement data useful for state and reward formulations. Studies commonly use available off-the-shelf microscopic simulation models including SUMO (Bálint et al. 2022; Bouktif et al. 2023; Li et al. 2020; Li et al. 2021; Liu et al. 2022; Long et al. 2022; Pang and Gao 2019; Shabestary et al. 2020; Shen et al. 2023; Yang and Fan 2024; Zhong et al. 2023), PTV Vissim (Cheng et al. 2022), Aimsun (Casas 2017; Hu et al. 2023) and Paramics (Li et al. 2016). In addition to allowing a realistic replication of traffic flow in a network, to implement RL, microscopic simulation engines should allow interaction during simulation, including state observation and signal control adjustment. For example, SUMO and PTV Vissim provide the Traci and component object module (COM) application programming interfaces (API), respectively (Lopez et al. 2018; PTV 2021). These interfaces allow access to most parameters during simulation. Additionally, to allow running of several hundred or thousands of simulations within a reasonable time frame, the selected simulation engine needs to provide a high level of run time efficiency.

### 2.3 RL-based TSP

A few recent studies have extended the RL-based signal control algorithms to include TSP. The common approach of incorporating TSP is to modify either one or both of the state and reward functions to include bus flow and its performance metrics (Cheng et al. 2022; Hu et al. 2023; Long et al. 2022; Shen et al. 2023; Yang and Fan 2024; Zhong et al. 2023). Except for Hu et al. (2023), the rest of the studies design second by second RL signal controllers incorporating bus parameters in the state and/or reward functions. The same algorithm is deployed to control general traffic and buses with bus entries populated with zeros when there are no buses on the approach. The majority of these studies inflate the bus arrivals/frequencies to generate the necessary samples for the agent to learn TSP control. Long et al. (2022) considers 12 to 60 buses/hour, Shen et al. (2023) considers 32 buses/hour, and Zhong et al. (2023) considers 37 buses/hour. Such frequencies create scenarios where a bus arrives within the time horizon of the disruption in the traffic caused by the TSP response to the previous bus. However, in the real world, bus arrivals are often scarce occurrences. Agents trained with high bus frequencies may be biased to the priority movements and may penalize nonpriority movements, such responses would be unnecessary in the absence of the high frequency bus arrivals. One means to address this challenge is to implement event-based TSP algorithms that are activated only in the presence of buses requiring TSP and are dormant otherwise.

## 3 METHODOLOGY

This section describes the formulation of the RL agents including the adapted simulation environment, the agent structure, selected hyper parameters, training, and testing.

### 3.1 Overview of Deep Q-network (DQN) and Double Deep Q-network (DDQN)

RL seeks to find the optimal mapping of states to actions, to achieve the highest numerical rewards. In DQN, a Q-value function is computed using a deep neural network (DNN). The DNN outputs Q-values corresponding to each action and the action with the highest q-value is selected. The model is trained on samples, also called experiences, stored in the memory at every time step. The stored experiences ($e_t$) at every time step consist of tuple of current state ($s_t$), action ($a_t$), next state ($s_{t+1}$), and reward ($r_{t+1}$), as shown in Equation (1). The stored experiences over time steps constitute what is termed as a memory buffer.

$$e_t = (s_t, a_t, s_{t+1}, r_{t+1}) \tag{1}$$

The training objective is to minimize the temporal difference error which is the difference between target Q-values and Q-values predicted by the model. Target values are approximated using the Bellman equation. During learning the algorithm tries to strike a balance between exploration and exploitation, commonly using decaying exploration probabilities. DDQN improves the stability of DQN by approximating the target Q-value with a separate network that is updated with the weights of the main network after a given number of episodes instead of every episode. Note that the term episode, which means a series of steps from start to finish, shall be used interchangeably with simulation run in the context of this study. Technical details of DQN can be found in Sutton and Barto (2018). Geron (2019) provides a good guide for implementation of the algorithm.

### 3.2 RL Agent Architecture

The conceptual architecture of the developed RL agents is shown in Figure 1. The architecture consists of a PTV Vissim simulation engine and two DDQN models, one for signal control of general traffic (DDQN-SC) in absence of a bus and one for implementing TSP (DDQN-TSP). First DDQN-SC is formulated and trained. It is then used as the background signal controller in the training of DDQN-TSP. During training of DDQN-TSP, and in the testing and implementation of the two trained algorithms, DDQN-SC controls the intersection until a bus requesting TSP enters the DSRC zone of the intersection. At this time control shifts to DDQN-TSP until the bus checks out. Figure 1 shows the training stage of DDQN-TSP. At the point of switching, the agent coming online takes the last state of the agent going offline. Over several episodes the agents learn to take actions that allow smooth transitions at these points.

Preliminary training was performed to select model hyperparameters. The final selected hyperparameters for both DDQN-SC and DDQN-TSP include: learning rate = 0.01, discount rate (gamma) = 0.99, exploration probability decay rate = 0.01, memory buffer capacity = 2000, neurons in two hidden layers = 64 and 128, and target network update frequency = 10 episodes. For brevity the process and results of hyperparameter tuning is not included in this paper. The state, reward, and actions shown in Figure 1 are discussed in detail later, after describing the simulation environment.

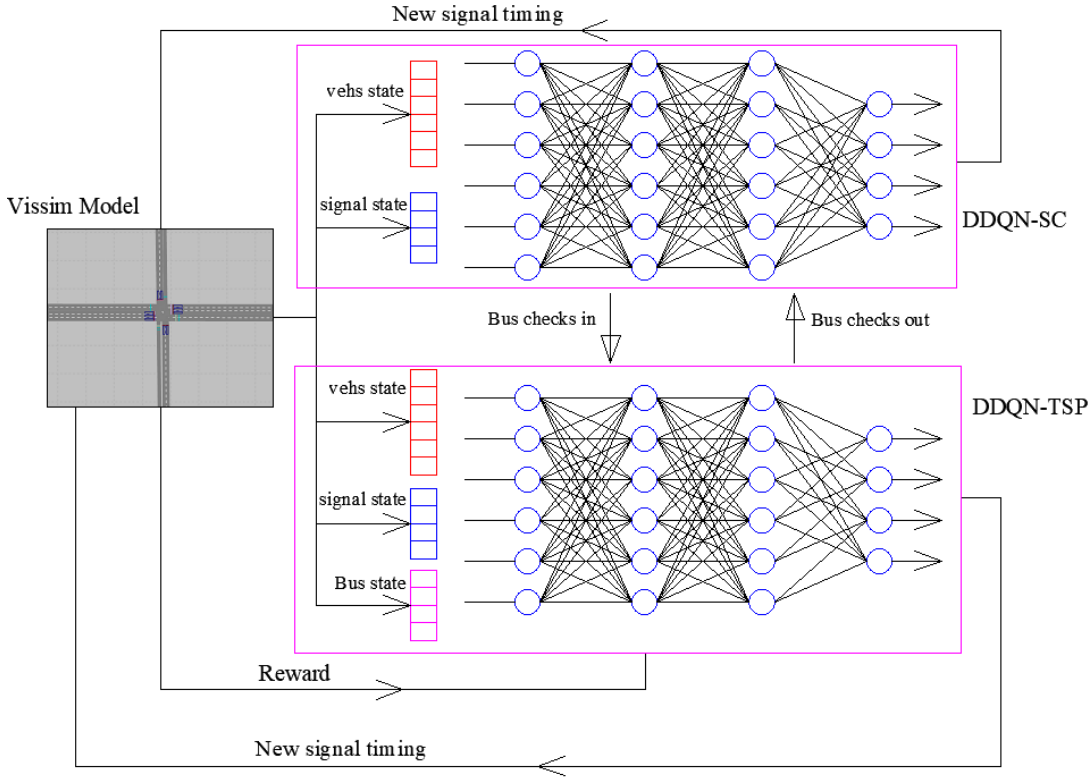

Figure 1: RL agent architecture during the training of DDQN-TSP.

## 3.3 Vissim Simulation Environment

Figure 2 provides details of the simulation environment and its interaction with the agent. PTV's Vissim, a widely used microscopic simulation platform, is selected as the simulation environment. As with other traffic microscopic simulation tools, PTV Vissim uses a combination of car following, lane changing, and gap acceptance models to model vehicle interactions and provides features to realistically represent road networks. The study utilizes a hypothetical single isolated intersection running actuated signal control in free mode (no fixed cycle). All approaches have stop bar detectors for fully actuated signal control. The network has a four-lane E-W major street and a two-lane minor street (N-S). All left turns on the major and minor streets have exclusive turn lanes and protected only signal phases. For simplicity, right turning movements are omitted from both streets. The model consists of a bus route on the main street (E-W), with a far side bus stop immediately downstream of the intersection. For model training, traffic volumes are selected to achieve a volume to capacity ratio (v/c) of approximately 0.95 when running fixed signal timings. Main street volumes are set to 1440 veh/h and 171 veh/h, for the through and left turn movements, respectively, while minor street volumes are 275 veh/h and 250 veh/h, for through and left turn movements, respectively. During the training of DDQN-TSP, buses enter the simulation with an average headway of 15 minutes (900 seconds) with a random term between -120 and +120 seconds, added to generate random arrivals. The 15 minutes is deemed sufficient to allow dissipation of the impacts of the prior TSP before another bus enters the system.

At each time step, the agent (either DDQN-SC or DDQN-TSP, whichever is active) provides new signal settings to the PTV Vissim intersection simulation, and the new state and reward value are computed with data extracted from the simulation and fed to the agent. PTV Vissim provides a COM interface, an API that enables interaction with the simulation in run time. However, running the simulation through COM can significantly increase the run time, especially if there is continuous interaction to exchange data with the simulation. As shall be seen in the results section, hundreds of episodes were required for each agent to learn the optimal policy. Thus run time efficiency becomes of great essence. This study adopts event-based scripts, a less widespread alternative to COM available in PTV Vissim. Event based scripting in Vissim involves embedding a script in the simulation and specifying functions to execute at given simulation time steps. In this study using event-based scripts was significantly faster than using COM.

However, using event-based scripts instead of COM required manually bridging model data, including 1) the training data (memory buffer) described earlier, 2) learned model parameters, that is, weights for both main and target network models, and 3) the exploration probability as it progressively decays. After each episode/simulation run the data is stored temporarily in a database and re-loaded at the start of the next simulation run. See the temporary storage module and its linkages in Figure 2.

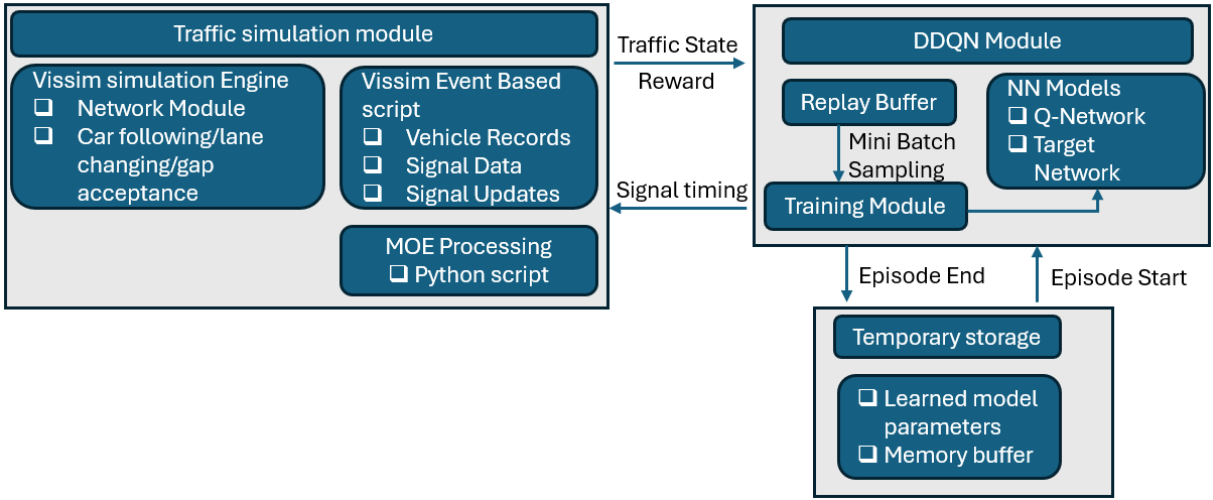

Figure 2: Traffic simulation and RL modules.

### 3.3.1 Simulation Execution

As the network is based on a hypothetical intersection, default model calibration was assumed, i.e., for car following the Wiedemann 74 model is utilized with default parameters. Lane change behavior, acceleration, etc., also utilize default parameters. Each episode lasts for 30 minutes during the training of DDQN-SC and 4 hours during the training of DDQN-TSP. More time is taken for DDQN-TSP to allow for sufficient bus samples at 15-minute headways. For both agents, random seeds remain unchanged during training. During testing, different random seeds are used for each run. Training on different random seeds showed slightly more instability and did not provide any meaningful benefits during testing compared to training on one random seed. Only results for one random seed training are included in the learning curves in the results section. Simulations runs and agent training are performed on an x64-based PC equipped with 12th Gen Intel(R) Core i9-12900, 2400 MHZ, 16 Core(s), 24 Logical Processor(s),128 GB of RAM, Intel (R) UHD Graphics 770 GPU and Windows 11 operating system.

## 3.4 State, Action and Reward Definitions

### 3.4.1 DDQN for general traffic control (DDQN-SC)

#### 3.4.1.1 State

As shown in Figure 1, the DDQN-SC state is defined by two vectors: (1) a vector of size 10, populated with the number of vehicles ("vehs state" in Figure 1) in each of 10 approach lanes, and (2) vector of size 4, populated with the green duration ("signal state" in Figure 1) for each signal phase. For the vehs state vector, the study assumes knowledge of the number of vehicles within 800 ft of the stop line for all inbound movements. The 800ft is assumed as the DRSC range. In the real world, the number of vehicles in each lane may not be readily available from the commonly deployed field sensors but may in future be available from CVs broadcasting their locations as they approach the intersection. Green duration refers to how long the current phase has been green and can be derived from Signal Phasing and Timing (SPaT) data.

#### 3.4.1.2 Action

The next phase is selected as the action. For simplicity four phases are defined, North/South Left, North/South Through, East/West Left, and East/West Through. All movements in a phase terminate at the same time. At every decision point (Δt of simulation time), the agent selects which phase to assign green. If the next phase is different from the current phase, a three second Yellow indication and one second Red are implemented to end the current Green and transition to the next phase. At the start of a phase, minimum green will be served. From the onset of yellow to the end of minimum green, the agent does not take any action. If the next phase selected is the same as the current phase, green time is extended by Δt. DQN-SC is trained and tested with Δt values of 1 second and 3 seconds. The training time was significantly higher for Δt = 1 compared with Δt = 3, and the training was less stable. Therefore Δt of 3 seconds was selected. For brevity the results with Δt = 1 are not included in this paper.

In addition to minimum green, the maximum green needs to be specified as well. Maximum green was computed by taking the fixed time green required for a v/c of 0.95, for each phase, and multiplying by 1.25 as recommended by FHWA's signal timing manual. In this study, an invalid action masking (IAM) algorithm is used to implement maximum green for each phase. When green duration of the current phase reaches maximum green, the predicted Q-value of the current phase is replaced with a large negative number to prohibit selection of the same phase.

#### 3.4.1.3 Reward

Base reward is defined as negative average delay for all vehicles. Negative of the delay is used in the reward function as delay is a disutility metric and the algorithm needs to move in the direction of positive or less negative rewards. As shown in Equation 2, average delay is obtained by summing delay for each vehicle ($d_i$) and dividing by total number of vehicles, n. To enable faster convergence, two penalty terms are introduced: (1) subtract a big number (N) from the base reward if queue length (ql) on any side street lane exceeds a set threshold ($ql_{Thr1}$) and (2) subtract a big number (M) from the reward if the agent switches phase from $\phi_t$ to $\phi_{t+1}$ when the movement currently receiving green still has a queue length ($ql_{\phi t}$) exceeding a set threshold ($ql_{Thr2}$). Training the agent with and without these penalties showed that the final model weights do not change significantly but with the penalties the models converge significantly faster.

$$reward = \begin{cases} -\frac{\sum_1^n d_i}{n}, & x < 0 \\ -\frac{\sum_1^n d_i}{n} - N, & ql_j > ql_{Th1} \\ -\frac{\sum_1^n d_i}{n} - M, & \phi_{t+1} \neq \phi_t, ql_{\phi t} > ql_{Th2} \end{cases} \quad (2)$$

### 3.4.2 DDQN for Transit signal priority (DDQN-TSP)

#### 3.4.2.1 State

The state definition for general traffic is expanded to include bus flow parameters as shown in Figure 1. In addition to the two vectors defined for DDQN-SC, two additional vectors are added for transit bus position and speed on the approach. It is assumed that all buses are connected, broadcasting their location and speed utilizing basic safety messages (BSM) as they approach the intersection. The DSRC range is taken as 800 ft and thus the bus only starts to communicate with the signal controller at 800 ft from the intersection. The 800 ft of the approach link is divided into 32 cells, each of 25 feet. The 25 ft cell size is selected to ensure that there is at most one bus in each cell at any given time step. Two vectors of length 32 with each entry representing a cell are created to represent bus position and speed. When the bus occupies a cell, the corresponding entry in the location matrix is populated with 1 and the with bus speed in the speed vector, otherwise both entries are populated with zero.

### 3.4.2.2 Action

Action for DDQN-TSP is also the selection of the next phase in the variable phasing sequence. Green extensions time step, ($\Delta t$) defined earlier remains set to 3 seconds.

### 3.4.2.3 Reward

Negative bus delay is taken as the reward and the same penalties are applied as in the DDQN-SC described above. Importantly, the penalty for side street queuing (N) and the side street queueing threshold are selected to balance tradeoffs between side street traffic delay and bus travel time.

## 4 RESULTS AND DISCUSSION

This section presents and discusses the study results. First the DDQN-SC agent is compared with actuated signal control (A-SC) to validate the agent's ability to generate signal timings that perform as well as well-timed current control strategies and qualify its use as a background controller in training DDQN-TSP. The second part of the results show the performance of the TSP agent including the learning progress during training and impacts on bus travel time and general traffic delay during the testing phase.

### 4.1 Performance of DDQN–SC

Figure 3 shows the learning curve for the DDQN-SC with the episode number on the x-axis and average reward for each episode on the y-axis. The average reward for an episode is computed by averaging rewards gained at each step in the episode. The algorithm progressively learns the best policy by exploration and exploitation, eventually converging after approximately 400 episodes. For the specified computer, each episode of 30 minutes requires on average 36.5 seconds for loading the model and all associated files and parameters, running the model for 30 simulation minutes, training the model at the end of episode, and saving the output. It is seen from the graph that stability improves (variability reduces) as the model converges. However, there is room for improving the stability, which may potentially be accomplished through other variations of DQN, including prioritized experience replay (PER), extended dueling, and others.

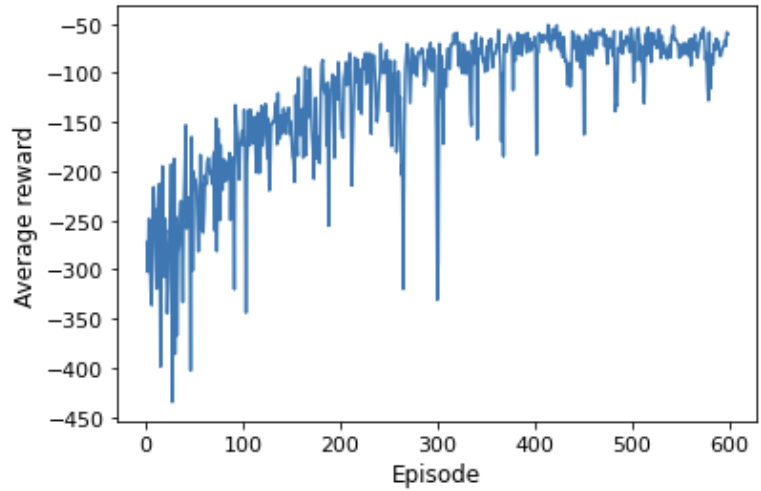

Figure 3: DDQN-SC learning curve.

Figure 4 shows a comparison of vehicle delay for four selected movements at the intersection for DDQN-SC and A-SC controls at (a) v/c =0.6 and (b) v/c= 0.95. A-SC uses the inbuilt RBC controller in PTV-Vissim running in free mode, with the same minimum and maximum green times as set for DDQN-SC. Eastbound through (EB_TH) is the main street through movement, southbound through (SB_TH) is the side street through movement, southbound left turn (SB_LT) is the side street left turn movement and eastbound left turn (EB_LT) is the main street left turn movement. The results are from 10 replicate runs, where each replicate is 1 hour long, with 15 minutes of warm up time and 45 minutes of data collection. The plotted data is the average vehicle delay from each replicate run and thus each box has 10 data points. The red square in the plot represents the average of the ten replicate run delays. For the side street movements (SB_TH and SB_LT), DDQN-SC results in lower delay compared to A-SC, with the difference more pronounced at v/c = 0.6. The main street through (EB_TH) delay is significantly lower for DDQN-SC at v/c = 0.6 and almost identical for both controls at v/c = 0.95. For main street LT (EB_LT), A-SC shows less delay at both levels of v/c, but again the difference is more evident at v/c = 0.6.

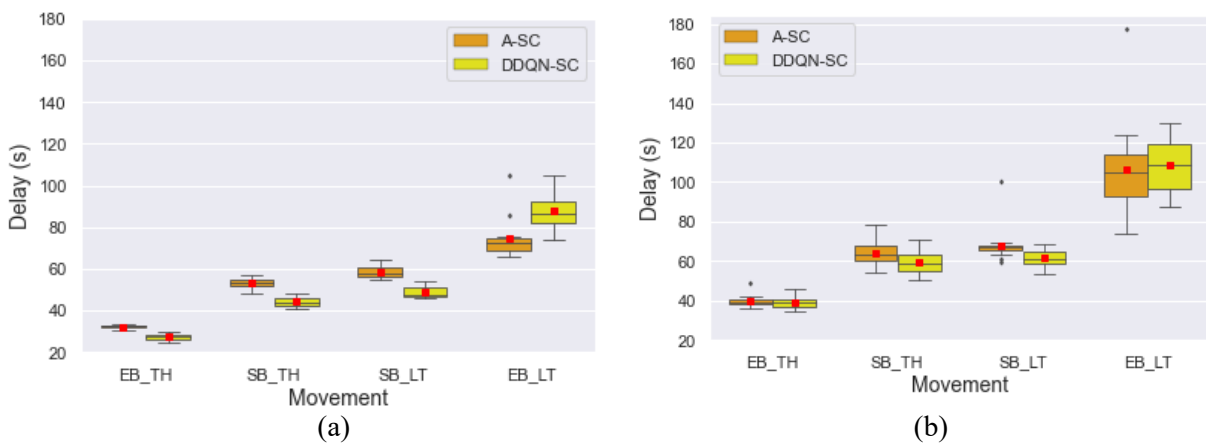

Figure 4: Comparing DDQN-SC and A-SC general traffic delay for (a) v/c = 0.6 and (b) v/c = 0.95

The observed relatively higher difference in performance of the two control types at v/c= 0.6 compared to v/c=0.95 is intuitively reasonable. At v/c= 0.95, the intersection is close to capacity and most movements consistently max out, resulting in operation close to fixed time control for both DDQN-SC and A-SC. At v/c = 0.6, where there is more flexibility for the optimization, DDQN-SC shows more benefits for movements with the highest volume. For example, on the main street, the EB_LT volume is 10% of the total approach volume and thus DDQN-SC favors EB_TH over EB_LT. Constraints in the reward function could be modified to alleviate this trade-off, if desired.

## 4.2  Performance of DDQN-TSP

Figure 5 (a) shows the learning curve for the DDQN-TSP with the episode number on the x-axis and average reward for each episode on the y-axis. As indicated in the methodology section, each episode is 4 hours long and includes 12 buses, with DDQN-TSP only running and collecting training data when the bus is on the approach. Figure 5 (b) shows the average bus delay during each episode. Each data point is an average of delay for the 12 buses in the episode. Bus delay progressively decreases with increasing episodes. From the figures it is seen that the algorithm converges after approximately 150 episodes.

Bus travel time with and without TSP at v/c =0.95 is shown in Figure 6 (a). For without TSP, DDQN-TSP is not invoked, and DDQN-SC provides signal timing while the bus is present. Each simulation run contains 12 buses and lasts for 4 hours. Bus headway is 15 minutes, with a random term between -120 and +120 seconds, added to generate random bus arrivals. Overall bus travel time is reduced by approximately 21% with DDQN-TSP. Additional benefits could be realized by extending the agent control from a single intersection to multiple intersections, which is a subject of an ongoing study.

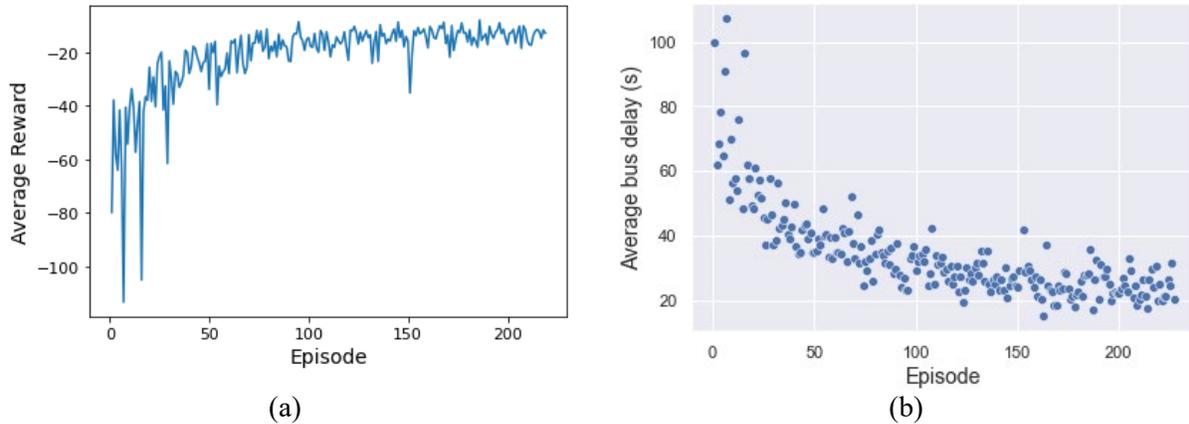

(a)                  (b)

Figure 5: (a) DDQN-TSP learning curve and (b) Average bus delay during training.

Figure 6 (b) shows the general traffic delay for 3 selected movements with and without TSP, for 10 simulation runs. The vehicles included in the analysis traverse the intersection in the time interval 5 minutes (300 seconds) after bus check-in. The interval of 300 seconds is chosen following a study by Guin et al. (2023) that showed that for v/c of 0.95, side street delay change persists up to about 300 seconds. It is seen that delay for the side street movements (SB_TH and SB_LT) marginally increases while the delay for main line through movement marginally reduces with TSP. This is intuitively reasonable as the main street through traffic benefits from increased green time given to the bus at the expense of side street traffic.

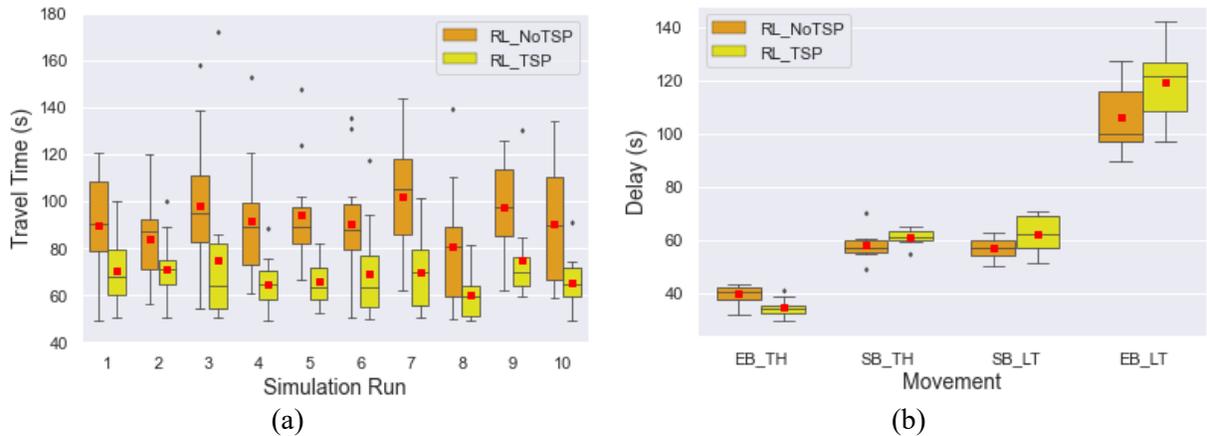

Figure 6: (a) Bus travel time with and without TSP and (b) General traffic delay with and without TSP

Lastly, a comparison is made between DDQN-TSP and A-SC TSP based on the bus travel time, Figure 7(a), and general traffic delay Figure 7(b). ASC_TSP and ASC_NoTSP, respectively, stand for A-SC control with and without TSP while RL_TSP and RL_NoTSP respectively stand for DDQN-TSP control with and without TSP. For TSP with A-SC, the inbuilt TSP algorithm in PTV Vissim's RBC is used with green extension, red truncation, and skipping of the conflicted phases allowed. Maximum green extension is set to 20 seconds. It is seen that DDQN-TSP performs slightly better in reducing bus travel time. For the side street impact, the two algorithms have very comparable performance. DDQN-TSP leads to a slightly greater decrease in main through movement (EB_TH) with TSP, which is consistent with providing more priority to the bus. The delay difference for SB_TH is almost the same for both algorithms while for SB_LT, A-SC seems to have less impact. Under the stated conditions, it is seen that the two algorithms show a very comparable performance, as evaluated from bus travel time and general traffic delay. Differences are likely to be seen if the algorithms are extended to multiple intersections which is a subject of an ongoing study.

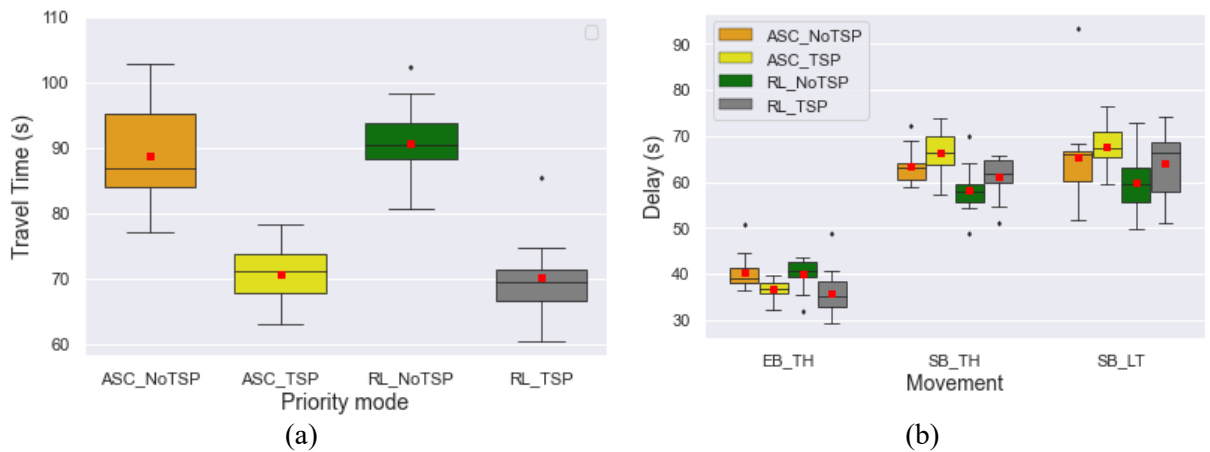

Figure 7: Comparison of DDQN-TSP and A-SC TSP based on (a) bus travel time and (b) General traffic delay.

## 5  CONCLUSIONS AND RECOMMENDATIONS

This study utilizes a microscopic simulation environment and CV data to develop and test an event-based RL agent that assumes intersection control from another RL-based traffic signal controller when TSP transit buses enter the dedicated short-range communication (DSRC) zone of the intersection. The background general traffic controller is trained and tested, demonstrating comparable performance with an actuated controller for a single intersection. The trained RL-based TSP agent is seen to reduce the bus travel time by about 21%, with marginal impacts to general traffic at a saturation rate of 0.95. The TSP agent also shows slightly better performance in improving bus travel time compared to actuated signal control with TSP. To improve run time efficiencies, PTV Vissim's event-based scripting is used instead of the commonly used COM API. Performance comparisons are limited to the traditional A-SC TSP systems, but this could be expanded to include other RL-based systems that have previously been proposed. In an ongoing study, the developed agents are being tested on multiple intersections in coordination. Additionally, software in the loop simulation with an emulator running the same software as field signal controllers will be used to further test the developed algorithms.

## AUTHOR BIOGRAHIES

**DICKNESS KAKITAHI KWESIGA** is a PhD student and a graduate research assistant in the School of Civil Engineering at Georgia Institute of Technology. His research interests include modeling, simulation and optimization of arterial corridor operations, traffic signal systems, Transit signal priority, emergency vehicle preemption, connected vehicles and AI/ML applications in traffic operations. His email is dkwesiga3@gatech.edu

**ANGSHUMAN GUIN** is a Senior Research Engineer in the School of Civil and Environmental Engineering at the Georgia Institute of Technology. His research attempts to find answers through innovations in the development of effective means of data collection, quality assurance, and processing to convert these data into informative metrics across a range of time-scales from near real time to decades. Dr. Guin's current research projects at Georgia Tech are broadly in Freeway Operations, Connected and Autonomous Vehicles, Intelligent Transportation Systems (ITS), Transportation Safety, Traffic Simulation and Data Management. His email is angshuman.guin@ce.gatech.edu

**MICHAEL HUNTER** is a Professor in the School of Civil and Environmental Engineering at Georgia Institute of Technology. His primary teaching and research interests are in transportation operations and design, specializing in adaptive signal control, traffic simulation, freeway geometric design, and arterial corridor operations. His email is michael.hunter@ce.gatech.edu